\documentclass[runningheads]{llncs}
\usepackage{graphicx}
\usepackage{booktabs}
\usepackage[utf8]{inputenc}
\usepackage[T1]{fontenc}
\usepackage{bm}
\usepackage{amsmath,amsfonts}
\usepackage{lineno}%
\usepackage{makecell}

\newcommand{\bi}{\begin{itemize}}
\newcommand{\ei}{\end{itemize}}
\newcommand{\ba}{\begin{array}}
\newcommand{\ea}{\end{array}}

\usepackage{xspace}

\newcommand{\brcksq}[1]{\left[#1\right]}

\newcommand{\eq}[1]{\begin{align}#1\end{align}}

\newcommand{\bmx}[0]{\begin{bmatrix}}
\newcommand{\emx}[0]{\end{bmatrix}}

\newif\ifarxiv
\arxivtrue

\begin{document}
\title{Evaluation of creating scoring opportunities for teammates in soccer via trajectory prediction}
%
\titlerunning{Evaluation of creating soccer scoring opportunities}
%

\ifarxiv
\author{Masakiyo Teranishi\inst{1} \and Kazushi Tsutsui\inst{1} \and Kazuya Takeda\inst{1} \and Keisuke Fujii\inst{1,2,3}}
%
\institute{Graduate School of Informatics, Nagoya University, Nagoya, Japan. \and
Center for Advanced Intelligence Project, RIKEN, Fukuoka, Japan.  \and 
PRESTO, Japan Science and Technology Agency, Saitama, Japan. 
\email{fujii@i.nagoya-u.ac.jp}
}
\authorrunning{M. Teranishi et al.}
\else
\author{Masakiyo Teranishi\inst{1} \and Kazushi Tsutsui\inst{1} \and Kazuya Takeda\inst{1} \and Keisuke Fujii\inst{1,2,3}}
%
\institute{Graduate School of Informatics, Nagoya University, Nagoya, Japan. \and
Center for Advanced Intelligence Project, RIKEN, Fukuoka, Japan.  \and 
PRESTO, Japan Science and Technology Agency, Saitama, Japan. 
}
\authorrunning{M. Teranishi et al.}
\vspace{-11pt}

\fi
\maketitle              
\begin{abstract}
\vspace{-11pt}
Evaluating the individual movements for teammates in soccer players is crucial for assessing teamwork, scouting, and fan engagement. 
It has been said that players in a 90-min game do not have the ball for about 87 minutes on average.
However, it has remained difficult to evaluate an attacking player without receiving the ball, and to reveal how movement contributes to the creation of scoring opportunities for teammates. In this paper, we evaluate players who create off-ball scoring opportunities by comparing actual movements with the reference movements generated via trajectory prediction. First, we predict the trajectories of players using a graph variational recurrent neural network that can accurately model the relationship between players and predict the long-term trajectory. Next, based on the difference in the modified off-ball evaluation index between the actual and the predicted trajectory as a reference, we evaluate how the actual movement contributes to scoring opportunity compared to the predicted movement. For verification, we examined the relationship with the annual salary, the goals, and the rating in the game by experts for all games of a team in a professional soccer league in a year. The results show that the annual salary and the proposed indicator correlated significantly, which could not be explained by the existing indicators and goals. Our results suggest the effectiveness of the proposed method as an indicator for a player without the ball to create a scoring chance for 
\ifarxiv
teammates.
\else
teammates\footnote{Appendix: \url{https://arxiv.org/abs/2206.01899}, *email: \url{fujii@i.nagoya-u.ac.jp}}.
\fi

\vspace{-6pt}
\end{abstract}
\vspace{-25pt}
\section{Introduction}
\vspace{-9pt}
\label{sec:introduction}
Assessing the movements of individual players for teammates in team sports is an important aspect of building teamwork, assessment of players' salaries, player recruitment, and scouting.
In soccer, most analytics has focused on the outcomes of discrete events near the ball (on-ball) \cite{Lucey2014,Decroos2017,Schulte2015,Schulte2017,Bransen2018,Decroos19,Liu2018,Liu2020} whereas much of the importance in player movements exist in the events without the ball (off-ball).
For example, it is said that players in a 90-min game do not have the ball for about 87 minutes on average \cite{Fernandez18}.
However, continuous off-ball movements are usually not discretized and difficult to understand except for core fans, experienced players, and coaches.
Also for the media and building fan engagement, quantitative evaluation of off-ball players is an issue in demand, which provides a common reference for beginners and experts in the sport e.g., when arguing a play of a favorite player.

Regarding the off-ball player evaluation methods, 
the positioning itself related to the goal was evaluated from the location data of all players and the ball. 
For example, the method called off-ball scoring opportunity (OBSO) to evaluate the player who receives the ball \cite{Spearman18} and the method to evaluate the movement to create space \cite{Fernandez18} have been proposed.
However, it has been still difficult to clarify how movements contribute to the creation of scoring opportunities for teammates, to evaluate other attacking players who do not receive it (e.g., a player moving tactically for teammates), and often to evaluate a score prediction to reflect the position of the multiple defenders. 

\begin{figure}[t]
    \centering
    \includegraphics[scale=0.3]{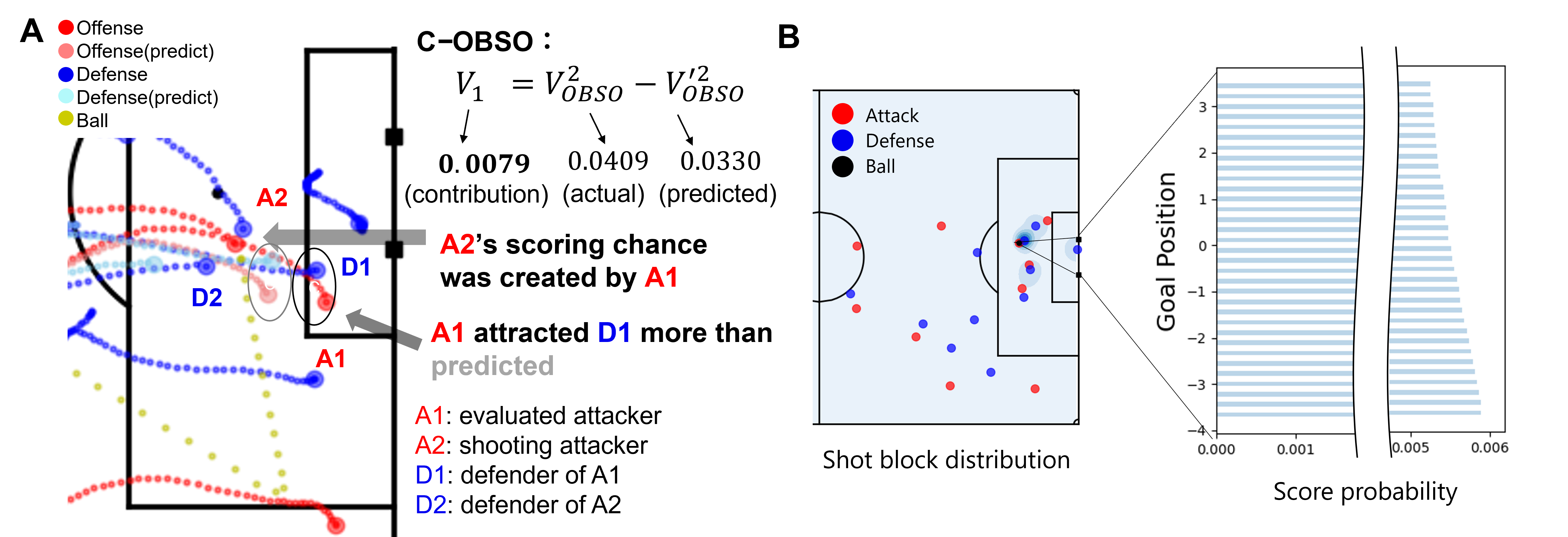}
    \vspace{-10pt}
    \caption{Our C-OBSO example and potential score model. (A) Example of C-OBSO computation. A1 is the player to be finally evaluated, A2 is the shooting player, D1 and D2 are the defender of A1 and A2, respectively. 
    $V_1$ is the C-OBSO value of A1, $V^2_{OBSO}$ is the actual OBSO, and $V'^2_{OBSO}$ is the reference OBSO value of A2 using the predicted trajectory.
    (B) Potential score model. Left: shot-blocking distribution formed by defenders. Right: shot probability corresponding to each shot vector.
    The vertical axis is the goal position (m), and the horizontal axis is the shot probability corresponding to each shot vector.
    }
    \label{fig:cobso_potential}
    \vspace{-15pt}
\end{figure}

In this paper, we propose a new evaluation indicator, Creating Off-Ball Scoring Opportunity (C-OBSO in Fig. \ref{fig:cobso_potential}A), aiming for evaluating players who create scoring opportunities when the attacking player is without the ball.
\ifarxiv
 The overview of our method is shown in Fig. \ref{fig:cobso_overview}.
\else
 The overview of our method is as follows.
\fi
(i) First, we modify the score model in the framework of OBSO \cite{Spearman18} with the potential score model that reflects the positions of multiple defenders with a mixed Gaussian distribution (Fig. \ref{fig:cobso_potential}B). 
(ii) Next, we accurately model the relationship between athletes and perform long-term trajectory predictions (Fig. \ref{fig:cobso_potential}A) using the graph variational recurrent neural network (GVRNN) \cite{Yeh2019}.
(iii) Finally, based on the difference in the modified off-ball evaluation index between the actual and the predicted trajectory (Fig. \ref{fig:cobso_potential}A), we evaluate how the actual movement contributes to scoring opportunity relative to the predicted movement as a reference.

In summary, our main contributions were as follows.
(1) We proposed an evaluation method of how movements contributed to the creation of scoring opportunities compared to the predicted movements of off-ball players in team sports attacks. 
(2) As a score predictor, we proposed a potential score model that considers the positions of multiple defenders in a mixed Gaussian distribution.
(3) In the experiment, we analyzed the relationship between the annual salary, the goals, and the game rating by experts, and show the effectiveness of the proposed method as an indicator for an off-ball player to create scoring opportunities for teammates.
Our approach can evaluate continuous movements of players by comparing with the reference (here predicted) movements, which are difficult to be discretized or labeled but crucial for teamwork, scouting, and fan engagement.
The structure of this paper is as follows.
First, we overview the related works in Section \ref{sec:related} and present experimental results in Section \ref{sec:experiment}.
Next, we describe our methods in Section \ref{sec:method} and conclude this paper in Section \ref{sec:conclusion}.

\vspace{-11pt}
\section{Proposed framework}
\vspace{-4pt}
\label{sec:method}

Here, we propose C-OBSO based on the motivation to evaluate players who create off-ball scoring opportunities for teammates.
To this end, in Section \ref{ssec:potential}, we first propose a potential score model that reflects the positions of multiple defenders with a mixed Gaussian distribution. 
Next, in Section \ref{ssec:c-obso}, we predict multi-agent trajectory using GVRNN \cite{Yeh2019}
and evaluate the difference between the actual value of the modified OBSO and the predicted value (as a reference) to evaluate how the movement contributed to the creation of scoring opportunities.

\vspace{-8pt}
\subsection{Potential score model in modified OBSO}
\vspace{-2pt}
\label{ssec:potential}
First, we describe the base model of our evaluation method called OBSO \cite{Spearman18} and then propose the potential score model.
OBSO evaluates off-ball players by computing the following joint probability
\vspace{-3pt}
\begin{align}
    P(G|D)&=\Sigma_{r\in R\times R}P(S_{r}\cap C_{r}\cap T_{r} |D) \\
    &=\Sigma_{r}P(S_{r}|C_{r}, T_{r}, D)P(C_{r}|T_{r},D)P(T_{r}|D),
\end{align}
where $D$ is the instantaneous state of the game (e.g., player positions and velocities).
\ifarxiv
 The details in OBSO are given in Appendix \ref{app:obso}.
\fi
$P(S_{r})$ is the probability of scoring from an arbitrary point $r\in R\times R$ on the pitch, assuming the next on-ball event occurs there.
$P(C_{r})$ is the probability that the passing team will control a ball at point $r$.
$P(T_{r})$ is the probability that the next on-ball event occurs at point $r$.
Here, for simplicity, we can assume that $P(S_{r}|D), P(T_{r}|D), P(C_{r}|D)$ are independent if the parameter $\alpha=0$ in the original work implementation (Eq. (6) in  \cite{Spearman18}).
Then, the joint probability can be decomposed into a series of conditional probabilities as follows:
\vspace{-3pt}
\begin{equation}
    P(G|D)=\Sigma_{r\in R\times R}P(S_{r}|D)P(C_{r}|D)P(T_{r}|D).
\end{equation}
$P(C_{r}|D)$ is the probability that the attacking team will control the ball at point $r$ assuming the next on-ball event occurs there, which is called the potential pitch control field (PPCF). 
$P(T_{r}|D)$ is defined as a two-dimensional Gaussian distribution with the current ball coordinates as the mean. 
$P(S_{r}|D)$ is simply calculated as a value that decreases with the distance from the goal. 
We used the grid data and computed $P(C_{r}|D)$ and $P(T_{r}|D)$ based on the code at \url{https://github.com/Friends-of-Tracking-Data-FoTD/LaurieOnTracking}.

In the original OBSO \cite{Spearman18}, the scoring probability was calculated as the output $P(S_{r}|D)$ of the score model as a function of the distance from the goal. 
However, the scoring probability may depend on the angle to the goal and the defensive position of the opponent. Therefore, in this paper, we propose a score model that reflects the angle to the goal and the position of multiple defenders.
Here, we consider the shot-blocking distribution of the defenders who can block shots in the field, and propose a potential model where the scoring probability decreases when defenders exist. 
The basic idea shown in Fig. \ref{fig:cobso_potential}B is to calculate the scoring probability from the angle to the goal at which the shot tends to be scored, considering the mixed distribution of the positions of multiple defenders.
The proposed scoring probability $P(S^p_{r}|D)$ at a certain point $r$ is calculated as the sum of the shot value $V_{shot}$ as follows:
\vspace{-6pt}
\begin{align}
    P(S^p_{r}|D) &= \sum_{i=1}^{n}V_{shot}(\vec{s_i}),\\
    V_{shot}(\vec{s}) &= C(c - V_{block}),
\end{align}
where $n$ is determined by the angle from the shooting position to the goal, and $\vec{s}$ is a shot vector per degree ($n$ is larger when the shot from the center and smaller from the side).
The shot value $V_{shot}$ is calculated by subtracting the shot block value $V_{block}$ from a certain constant $c$
($c, C$ are parameters determined from data to be adjusted so that $V_{shot} \geq 0$ and $P(S^p_{r}) \in [0,1]$).
Let $V_{block}$ be the sum of the shot block distribution values along the shot vector $\vec{s}$.
The shot blocking distribution is the sum of the normal distributions (variance $\sigma^2=0.5+l_d$) assigned to each defender on the goal side of the shooting position (shot blockable players using legs), where $l_d$ is the distance between the shooting position and the defender.
We consider that goalkeepers have a shot blocking distribution with twice the value of normal defenders because of a higher shot-blocking ability.
Here, we assume that the block distribution is not changed with the distance from the ball.
A defender near to the ball may affect the ball, but far players use the flight time of the ball for their movement.
This formulation is left for future work. 
 
\vspace{-8pt}
\subsection{C-OBSO with trajectory prediction}
\vspace{-3pt}
\label{ssec:c-obso}
Here, we describe the base model of our trajectory prediction method called GVRNN \cite{Yeh2019} and then describe our C-OBSO framework.
Our contribution here is to evaluate how the actual ``off-ball'' movement contributes to scoring opportunity compared to the predicted movement (or trajectory) as a reference. 
In our method, we use GVRNN \cite{Yeh2019}, which is a VRNN \cite{Chung15} combined with a graph neural network (GNN \cite{Kipf18}). 
\ifarxiv
 For the details in VRNN and GVRNN, see Appendices \ref{app:vrnn} and \ref{app:GVRNN}.
\fi
In GVRNN, the graph encoder-decoder network models the relationship between players as a graph, which is one of the best performing models for predicting player trajectories in team sports \cite{Yeh2019}. 
This is a probabilistic model which can sample multiple possible trajectories.

Based on the trajectory prediction, we propose an evaluation index C-OBSO of players who create scoring opportunities for teammates. 
The basic idea is to evaluate an off-ball player from the difference in the modified OBSO values between the predicted and actual movements of the players.
The C-OBSO value of a player $i$ without the ball can be expressed as follows.
\vspace{-3pt}
\begin{equation}
    V_i = V_{OBSO}^k - V'^k_{OBSO}~,
\end{equation}
where the player $k$ is the ball carrier who performs a final action (e.g., shot), $V^{k}_{OBSO}$ is the modified OBSO in the actual game situation, and $V'^{k}_{OBSO}$ is the modified OBSO based on the predicted trajectory as a reference.
For example, in Fig. \ref{fig:cobso_potential}A, C-OBSO is positive and the player to be evaluated (A1) contributes more to the shooter (A2) than the referenced (predicted) player. 
Specifically, A1 has created a more advantageous situation for A2 by attracting D1 more than expected. 
C-OBSO can evaluate a player in such situations with an interpretable value (i.e., the increase in scoring probability).
Theoretically, if perfectly predicted, C-OBSO is always zero, but actually, if we apply this to a test data, the perfect prediction is impossible. 
In other words, we assume the imperfect trajectory prediction in this framework.

\vspace{-12pt}
\section{Experiments}
\vspace{-6pt}
\label{sec:experiment}
In this section, we validate the proposed method of the potential score model, the trajectory prediction model (GVRNN), and the C-OBSO itself.
For our implementation, the code is available at \url{https://github.com/keisuke198619/C-OBSO}. 

\vspace{-11pt}
\subsection{Dataset}
\vspace{-6pt}
\label{ssec:dataset}
In this study, we used all 34 games data of Yokohama F Marinos in the Meiji J1 League 2019 season to perform specific player-level evaluations in limited data.  
Note that the tracking data for all players and timesteps were not publicly shared in such amounts. 
The dataset includes event data (i.e., labels of actions, e.g., passing and shooting, recorded at 30 Hz and the simultaneous xy coordinates of the ball) and tracking data (i.e., xy coordinates of all players recorded at 25 Hz) provided by Data Stadium Inc.
The company was licensed to acquire this data and sell it to third parties, and it was guaranteed that the use of the data would not infringe on any rights of the players or teams.
For annual salaries, we used the salaries of the same team (Yokohama) in 2019 \cite{SoccerMoney} because they were different valuation criteria for different teams and the transfer of the players took place during the season.
The goals for each player in each match were collected from \cite{Score}.
The rating by experts in each match \cite{Rating} was also used for verification, which was scored in 0.5 point increments with a maximum of 10 points.

\vspace{-11pt}
\subsection{Data processing for verification}
\vspace{-6pt}
\label{ssec:process}
We used the attacking data of Yokohama F Marinos for the test and those of the opponent teams for training the model or parameter fitting. 
Again, since the data was limited in this study, we split the data in such a way, and if we have more data, we can analyze all teams with the training data with the same team. 
Here we describe the processing of the potential score model, the trajectory prediction model, the C-OBSO, and their statistical analyses.

\vspace{-11pt}
\subsubsection{Potential score model.}
To validate the potential score model, the opponent's shots (345 shots, 34 goals) were used for fitting the parameters $c$ and $C$, and Yokohama F Marinos' shots (494 shots, 59 goals) were used for verification.
The parameters $c, C$ of the potential score model were determined to be $c = 1.1,~C = 1/150$ using the data of the opponents. 
The potential score model was verified by the root mean square error (RMSE) between the actual score and the calculated scoring probability.
We compared the RMSE with that of a simple score model as a function of distance from the goal for implementing the original OBSO \cite{Spearman18} (see also Section \ref{ssec:potential}).
Although there have been more holistic score models such as \cite{Fernandez2019,anzer2021goal}, to fairly compare with our potential model as a component of the modified OBSO, we consider the simple score model as an appropriate baseline.

\vspace{-11pt}
\subsubsection{Trajectory prediction model.}
For the test data of trajectory prediction and C-OBSO, we used 412 shot scenes of Yokohama F Marinos (we selected the sequences of consecutive events and excluded too short events such as a free kick).
The trajectory prediction model was trained using the opponents' data to generate ``league average'' trajectories.
The tracking data were down-sampled to 10 Hz (after prediction, up-sampled at the original 25 Hz) based on \cite{fujii2020policy}.
To verify the accuracy of the long-term trajectory prediction, we set various time lengths (6, 8, 10, and 12 s) using mean trajectories in 10 samples. 
We divided into the opponent data for batch training (6 s: 94208 sequences, 8 s: 49152 sequences, 10 s: 33536 sequences, 12 s: 24320 sequences) and the validation (6 s: 10477 sequences, 8 s: 5479 sequences, 10 s: 3730 sequences, 12 s: 2721 sequences). 
Note that the end of all sequences was the moment of a shot. 
The input feature has 92 dimensions (the xy coordinates and the velocity of 22 players and the ball). 
During training, the model was trained based on the one-step prediction error of all combinations of the two attackers who invaded the attacking third.
We simultaneously predicted the three players: one of the off-ball attackers and the defenders closest to each attacker. 
Note that we only consider the three players' interactions and ignore others' interactions, 
because the prediction error will increase if the numbers increase, and the increase of the predicted players is left for future work.

For the test data of the 412 sequences, the three relevant players and the attacker from the same criterion were predicted. 
At the inference, using 2 s sequences as burn-in period, we predicted the sequences for the subsequent time lengths (i.e., 4, 6, 8, and 10 s) by updating the estimated position and velocity (i.e., performed long-term prediction). 
For the training of the proposed and baseline models, we used the Adam optimizer \cite{Kingma15} with a learning rate of $0.001$ and $10$ training epochs.
We set the batchsize to 256.
For the performance metrics, we used the endpoint error (mean absolute error: MAE) from the actual trajectory. 

\vspace{-15pt}
\subsubsection{C-OBSO.}
To compute C-OBSO, predicted trajectories with 4 s (total 6 s) were used. 
This is because a longer prediction time will result in a larger prediction error, while a shorter prediction time will not make a difference in the evaluation of C-OBSO.
Although the negative values of C-OBSO are also calculated by comparison with the reference, the negative values were calculated as 0, assuming that they may not have a negative effect on the behavioral players. 
This is because there were many situations with negative values in which the shooter's defender did not take an appropriate defensive position in the predicted trajectory.

\vspace{-11pt}
\subsubsection{Statistical analysis.}
For the verification of C-OBSO, we examined the relationship with the annual salary, the goals, and the expert's rating. 
Note that there is no ground truth available for the verification.
We also compared them with the existing OBSO \cite{Spearman18}. 
Since some of the data often did not follow normal distributions, we used Spearman’s rank correlation coefficient $\rho$ for these relationships. 
Regarding the RMSE in the potential score model and MAE in the trajectory prediction, for the same reason, we used nonparametric statistical tests to compare with the baselines.
Regarding the potential score model, we used the Wilcoxon rank sum test. 
For all statistical calculations, $p < 0.05$ was considered as significant. 
\vspace{-11pt}
\subsection{Our model verification}
\vspace{-6pt}
\label{ssec:verify-psm}
First, we validated the potential score model needed to calculate the C-OBSO.
The RMSE with the actual scores was 0.324 $\pm$ 0.014 for the conventional score model \cite{Spearman18} without considering the defenders and goal angles, and 0.309 $\pm$ 0.0014 for the potential score model ($p < 10^{-10}$). 
This result suggests that the proposed method models the scores more accurately.

\begin{figure}[h]
    \centering
    \includegraphics[scale=0.45]{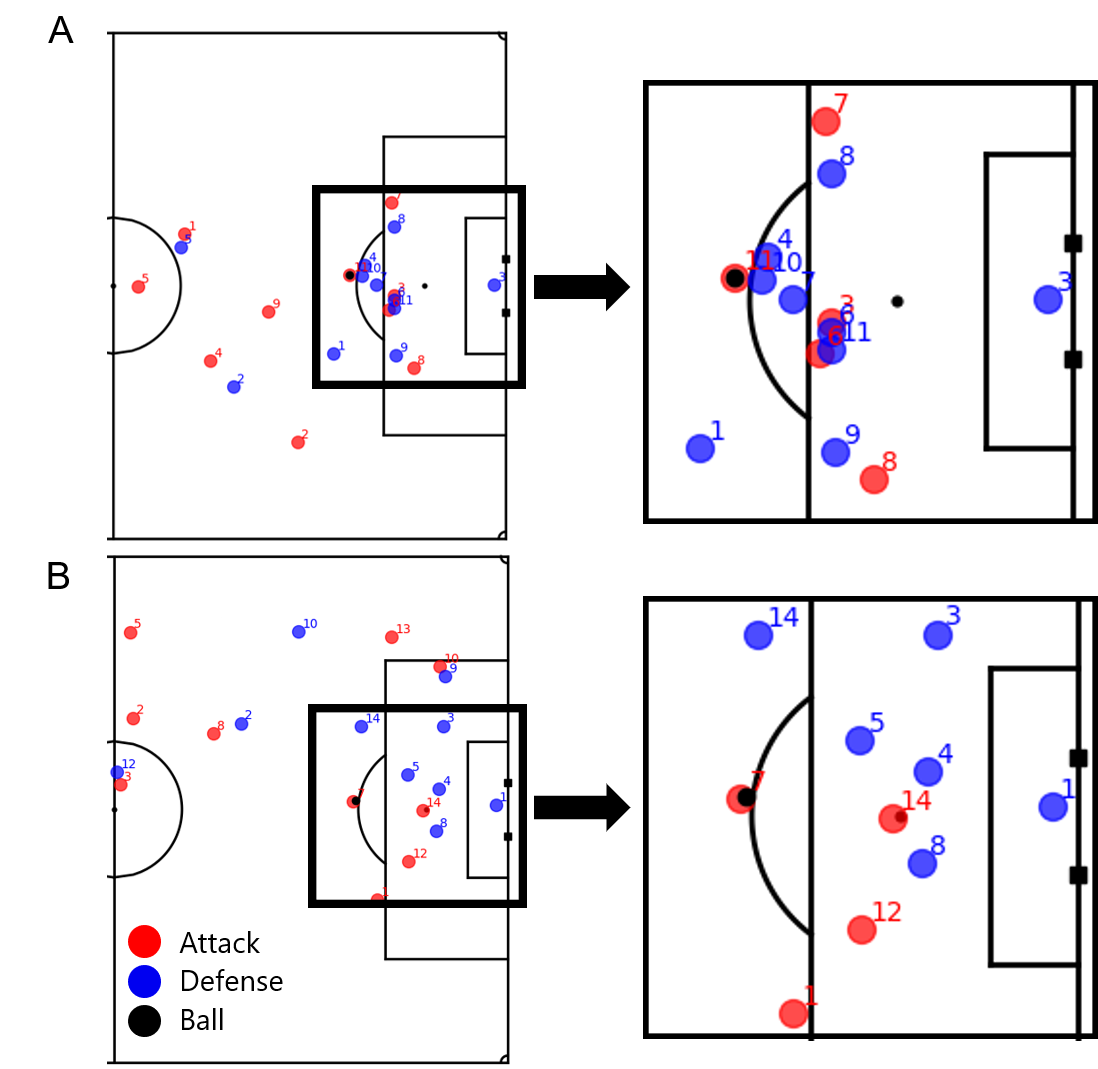}
    \caption{
    Comparison between the score model of conventional and the proposed potential model. 
    The scoring probability of our model is lower (A)  when the defenders are crowded than (B), whereas that of the conventional score model in (A) was the same as (B).
    }
    \label{fig:shot_compared}
    \vspace{-10pt}
\end{figure}

Figure \ref{fig:shot_compared} shows an example of the two methods in two actual situations where a shot is attempted from a similar distance. 
In the existing method, the probabilities were the same (both 0.1237) because the shots were taken from almost the same distance. 
The proposed method had a lower scoring probability with more defenders (upper: 0.0489, lower: 0.1202).
We indicate that the proposed method reflects the position of multiple defenders and can model the score accurately.

Next, we show the results of the trajectory prediction model for computing C-OBSO.
Endpoint errors (MAE and standard error, [m]) in GVRNN were $0.608 \pm 0.014, 0.867 \pm 0.022, 1.701 \pm 0.045, 1.606 \pm 0.042$ in 4, 6, 8, 10 s prediction.
In GVRNN, longer predictions show larger prediction errors except for the difference between 8 s and 10 s.
Since the 4 s prediction of GVRNN achieved a low the MAE of less than 0.7 m, the GVRNN trajectory prediction of 4 s was used in the next C-OBSO.
\ifarxiv
For details, see also Appendix \ref{app:prediction_results}.
\fi


\begin{figure*}[t]
    \centering
    \includegraphics[scale=0.32]{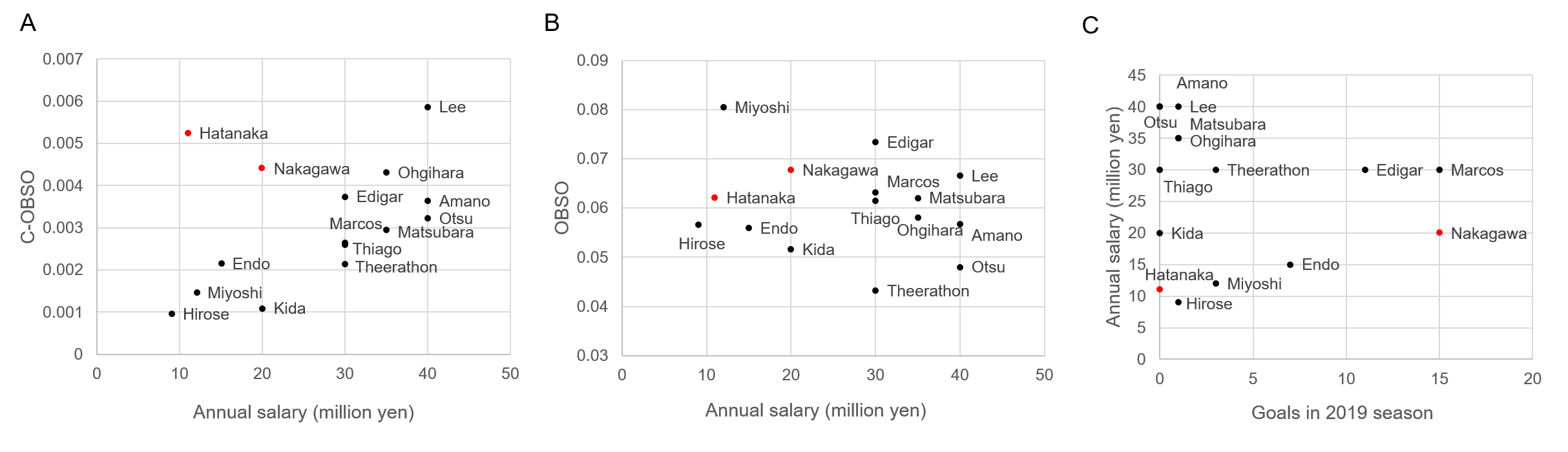}
    \vspace{-10pt}
    \caption{Relationship between indicators, goal, and annual salary in a team. 
    (A) Relationship between C-OBSO and the salary. 
    (B) Relationship between OBSO \cite{Spearman18} and the salary. 
    (C) Relationship between each player's goals and annual salary. 
    Red players received individual awards (Hatanaka: valuable player Award, Nakagawa: the most valuable player award). 
    }
    \label{fig:salary_cobso}
    \vspace{-10pt}
\end{figure*}

\begin{figure*}[h]
    \centering
    \includegraphics[scale=0.3]{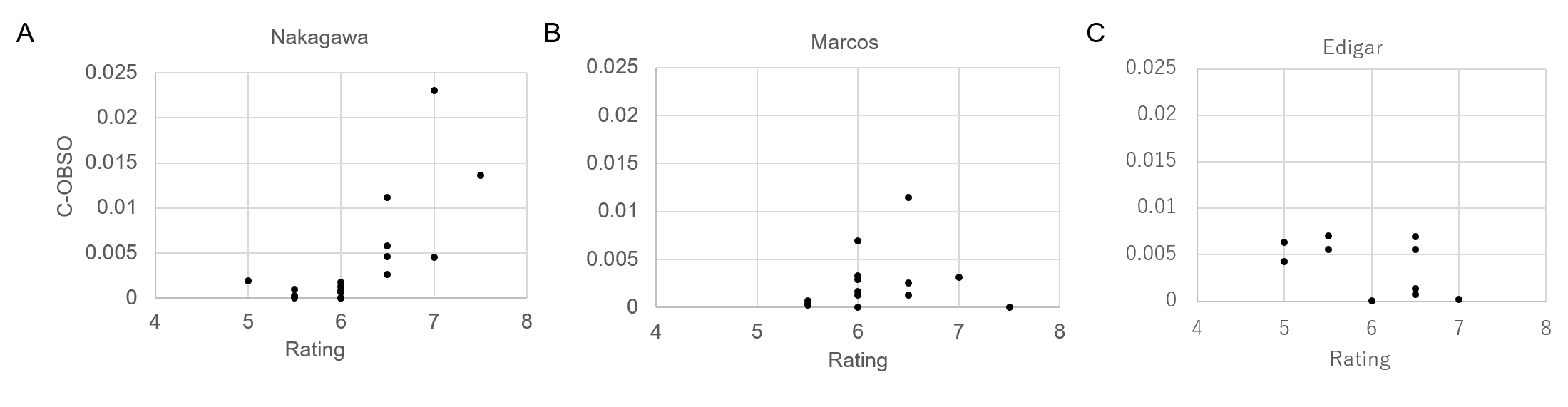}
    \vspace{-10pt}
    \caption{Relationship between C-OBSO and the rating by experts of the top three scorers (A: Nakagawa, B: Marcos, C: Edigar) for each game.
    }
    \label{fig:cobso_scoreplayer}
\end{figure*}

\begin{figure*}[h]
    \centering
    \includegraphics[scale=0.32]{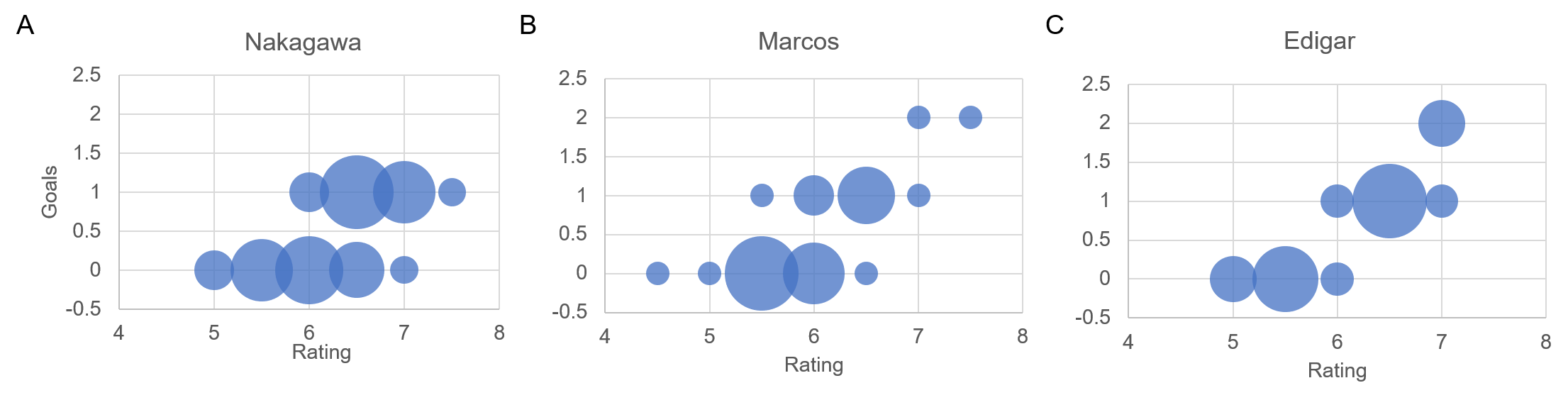}
    \vspace{-10pt}
    \caption{Relationship between the goals and the rating by experts of the top three scorers (A: Nakagawa, B: Marcos, C: Edigar) for each game.
    The size of the circle represents the frequency because there are many combinations of the goals and the rating with the same value.
    }
    \label{fig:rating_scoreplayer}
\end{figure*}

\vspace{-11pt}
\subsection{C-OBSO results}
\vspace{-4pt}
\label{ssec:verify-c-obso}
Verification of C-OBSO is challenging because of no ground truth values or player ratings. Therefore, we analyzed the relationship with the annual salary, the goals, and the game rating by experts, whereas we admit that these variables include various confounding factors.
The relationships between the average C-OBSO and OBSO values of each player of Yokohama F Marinos in 2019 and the annual salary of each player in 2019 are shown in Fig. \ref{fig:salary_cobso} (note that the tracking data for all players and timesteps were not publicly shared). 
Here we analyzed 15 players with more than 10 sequences under evaluation.
As a result, there was a significant positive correlation between annual salary and C-OBSO ($\rho = 0.45, p = 0.046$). 
In addition, the two players with the higher evaluation values but lower salaries (in red in Fig. \ref{fig:salary_cobso}A) were highly evaluated players, who won the individual awards (the most valuable player and valuable player award). 
In fact, their annual salary for the following year (2020) was also increased (valuable player: increased from 11 million yen to 40 million yen; the most valuable player: increased from 20 million yen to 60 million yen).

We found that these tendencies were similar to the C-OBSO and OBSO without the potential score 
\ifarxiv
model (see Appendix \ref{app:withoutpotential}).
\else
model.
\fi
On the other hand, there was no significant correlation ($\rho=-0.28, p=0.154$) for OBSO, which evaluates a player's own scoring opportunities (Fig. \ref{fig:salary_cobso}B).
We also examined the relationship between annual salary and goals (Fig. \ref{fig:salary_cobso}C), and found no significant correlation ($\rho=-0.23, p=0.208$). 
Therefore, there was no relationship between annual salary and goals. 
There were many players with zero goals, and it is difficult to evaluate them only with the goals.

Next, in order to examine the relationship with player performance in more detail, we show the relationship between C-OBSO and the rating by experts of the top three scorers (Nakagawa with 15 goals, Marcos with 15 goals, and Edigar with 11 goals in this season) in Fig. \ref{fig:cobso_scoreplayer}.
We analyzed the games in which there were two or more C-OBSO evaluations using the average of C-OBSO values on each game (17 games for Nakagawa, 14 games for Marcos, and 10 games for Edigar).
A strong positive correlation was found only for Nakagawa ($\rho=0.75,p=0.0003$) but not for Marcos ($\rho=0.27,p=0.174$) and Edigar ($\rho=-0.37,p=0.145$).
\ifarxiv 
 In Appendix \ref{app:rating}, we show the results of the other four players who played seven games or more and had two related scoring opportunities or more (for C-OBSO). 
\fi
Similarly, there were no significant correlations between them for all players ($\rho s < 0.190,~ ps > 0.05$). 
In addition to the number of scoring opportunities for teammates (17 times), the results found that Nakagawa would be subjectively and quantitatively an outstanding player.

For reference, we also show the relationship between the goals of the top three scorers and the ratings by experts in Fig. \ref{fig:rating_scoreplayer}.
We analyzed the games in which each player played (33 games for Nakagawa, 33 for Marcos, and 16 for Edigar). 
For each player, there were strong correlations between the goals and the rating (Nakagawa $\rho=0.63,p=4.33\times 10^{-5}$, Marcos $\rho=0.71,p=1.98\times 10^{-6}$, Edigar $\rho=0.91,p=4.40\times 10^{-7}$). 
We found that the rating of each game depends on a rare event (i.e., goals).
\ifarxiv 
 In Appendix \ref{app:rating}, we show the results of the other four players who scored two points or more.
\fi
Similarly, there were significant correlations between them for all players ($\rho s > 0.516,~ ps < 0.018$). 
Recall that there was a stronger correlation between C-OBSO and Nakagawa's rating than for the other two players. Nakagawa also had higher average ratings than the other players (6.26 for Nakagawa, 5.97 for Marcos, and 6.09 for Edigar), and he was the player who won the most valuable player award. 
The game rating by experts would depend on the goals, but it may also evaluate the creation of scoring opportunities only for Nakagawa. 
From these results, we speculate that Nakagawa was highly evaluated not only for his scoring but also for his contribution to other attacking players.
Our method can also evaluate players difficult to be evaluated by conventional indicators, which is crucial for assessing teamwork and player salary, player recruitment, and scouting.

\vspace{-10pt}
\section{Related work}
\vspace{-10pt}
\label{sec:related}
In the tactical behaviors of team sports, 
agents select an action that follows a policy (or strategy) in a state, receives a reward from the environment and others, and updates the state \cite{fujii2021data}. 
This is similar to a reinforcement learning framework (e.g., \cite{Bernstein02}). 
Due to the difficulty in modeling the entire framework from data for various reasons \cite{van2021learning} (e.g., a sparse reward and difficulty in estimating intents), we can adopt two approaches: to estimate the related variables and functions from data (i.e., inverse approach) as a sub-problem, and to build a model (e.g., reinforcement learning model) to generate data in virtual space (i.e., forward approach, e.g., \cite{kurach2020google,scott2022does}). 
Here, we focus on the former approach and introduce the research from the view of inverse approaches.

There have been many approaches to quantitatively evaluate the actions of attacking players about the scoring, such as based on the expected scores using tracking data \cite{Lucey2014,Decroos2017,Schulte2015,Schulte2017,Bransen2018}, 
action data such as dribbling and passing \cite{Decroos19,dick2021rating}, and estimating state-action value function (Q-function) \cite{Wang2018,Liu2018,Liu2020}.
Some researchers have evaluated passes \cite{Power17,Brooks2016,dick2022can}, and others evaluated actions to receive a ball by assigning a value to the location with the highest expected score \cite{Spearman18,Link2016} and a rule-based manner \cite{fujii2020cognition}.
In particular, Spearman \cite{Spearman18} proposed an evaluation metric called OBSO to evaluate behavior based on location data and rule-based modeling.
Defensive behaviors have also been evaluated based on data-driven \cite{Robberechts2019,Toda2021} and rule-based manners (e.g., \cite{Teranishi2020}).
However, these score evaluations do not often reflect the position of multiple defenders and goal angles in rule-based manner.

From the perspective of reinforcement learning,
there have been many studies on inverse approaches.
As for the study of state evaluation, there are several studies based on score expectation (e.g., \cite{Cervone2014,Cervone2016ASA,Fernandez2019}) and based on the value of space  (e.g., \cite{Cervone2016SSAC,Fernandez18}).
There is also research on estimating reward functions by inverse reinforcement learning \cite{Luo2020,Rahimian2020}.
Researchers performed trajectory prediction sometimes in terms of the policy function estimation, as imitation learning \cite{Le17,le2017data,Teranishi2020,fujii2020policy} and behavioral modeling \cite{Zhan19,Yeh2019,Li2021,fujii2022estimating} to mimic (not optimize) the policy using neural network approaches.
In this paper, we first propose a method to evaluate how the actual ``off-ball'' movement contributes to scoring opportunity 
based on the difference between the state values generated from the actual and the reference policies.



\vspace{-11pt}
\section{Conclusion}
\vspace{-6pt}
\label{sec:conclusion}
In this paper, we evaluated players who create off-ball scoring opportunities by comparing actual movements with the reference movements generated by trajectory prediction. 
Our results suggest the effectiveness of the proposed method as an indicator for a player without the ball to create scoring opportunities for teammates.
For future work, although the number of players to be evaluated was determined in the minimum setting, it is possible to evaluate the contribution to the scoring opportunities for teammates in a less limited way by predicting a larger number of players in both offense and defense.
Furthermore, since our method evaluates off-ball players by comparing them with the referenced trajectory, the value becomes too small. 
Computing the evaluation value in a more interpretable way (e.g., in a score scale) would be future work.
Finally, computing our indicators from broadcast videos (e.g., \cite{deliege2021soccernet}) or other videos (e.g., top- or side-view \cite{scott2022soccertrack}) would also be future work.

\vspace{-10pt}
\section*{Acknowledgments}
\vspace{-5pt}
This work was supported by JSPS KAKENHI (Grant Numbers 20H04075 and 21H05300) and JST Presto (Grant Number JPMJPR20CA).

\bibliographystyle{splncs04}
\ifarxiv

\else
\bibliography{reference}
\fi

\ifarxiv
 \newpage
 \renewcommand{\thesection}{\Alph{section}}
\appendix

\section*{Appendix}
\section{Overview of our method}
\label{app:overview}
The overview of our method is shown in the Fig. \ref{fig:cobso_overview}.
(i) First, we modify the score model in the framework of OBSO \cite{Spearman18} with the potential score model that reflects the positions of multiple defenders with a mixed Gaussian distribution (Fig. \ref{fig:cobso_potential}B). 
(ii) Next, we accurately model the relationship between athletes and perform long-term trajectory predictions (Fig. \ref{fig:cobso_potential}A) using the graph variational recurrent neural network (GVRNN) \cite{Yeh2019}.
(iii) Finally, based on the difference in the modified off-ball evaluation index between the actual and the predicted trajectory (Fig. \ref{fig:cobso_potential}A), we evaluate how the actual movement contributes to scoring opportunity relative to the predicted movement as a reference. 

\begin{figure*}[h]
    \centering
    \includegraphics[scale = 0.55]
    {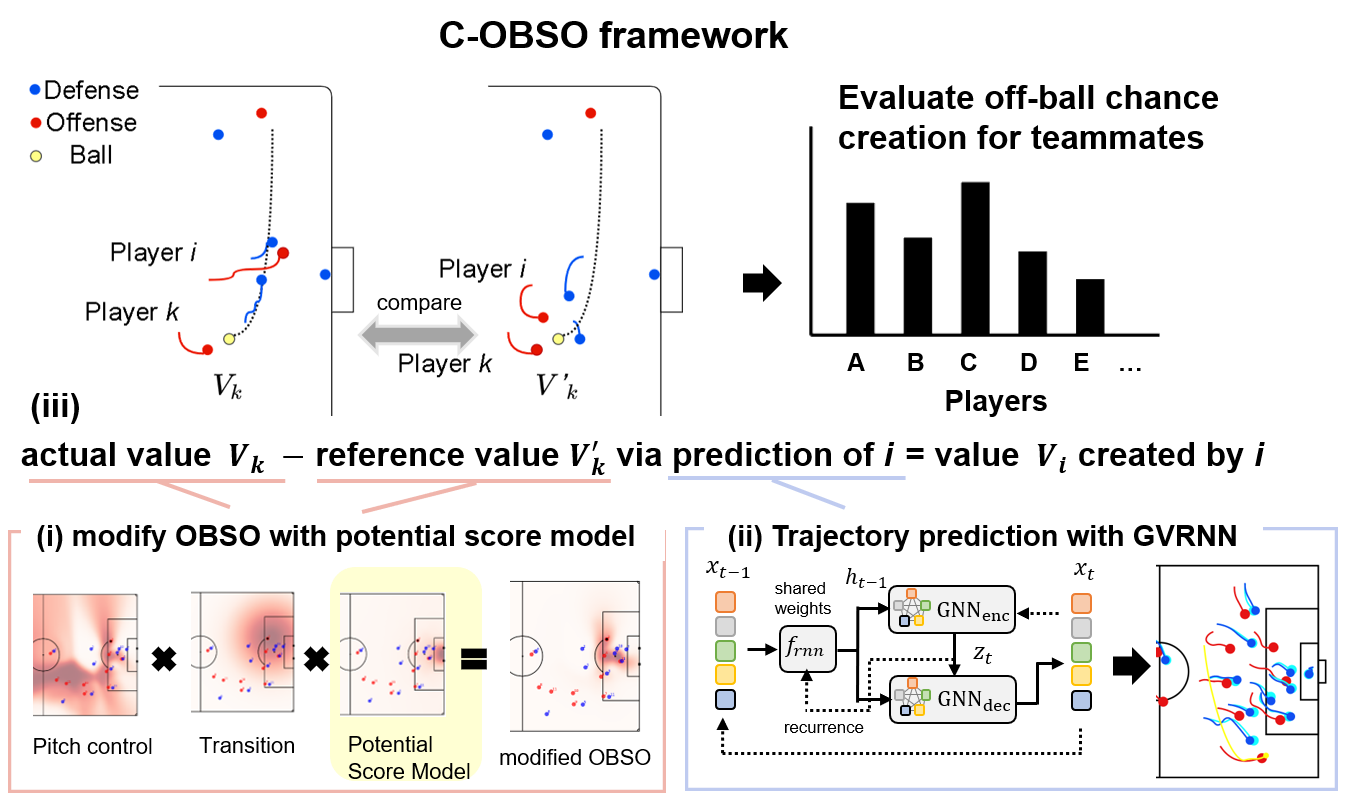}
    \caption {Overview of our method. (i) First, we propose a potential score model to improve OBSO \cite{Spearman18}. (ii) We then predict players' trajectories using GVRNN \cite{Yeh2019} to generate a reference player trajectory. (iii) Finally, the proposed C-OBSO is calculated by the difference between the evaluation value in the actual game situation and the referenced or predicted game situation.}
    \label{fig:cobso_overview}
\end{figure*}

\section{Off-ball scoring opportunity \cite{Spearman18}}
\label{app:obso}
Here, we describe the base model of our evaluation method called OBSO \cite{Spearman18}.
OBSO evaluates off-ball players by computing the following joint probability
\vspace{-3pt}
\begin{align}
    P(G|D)&=\Sigma_{r\in R\times R}P(S_{r}\cap C_{r}\cap T_{r} |D) \\
    &=\Sigma_{r}P(S_{r}|C_{r}, T_{r}, D)P(C_{r}|T_{r},D)P(T_{r}|D),
\end{align}
where $D$ is the instantaneous state of the game (e.g., player positions and velocities).
$P(S_{r})$ is the probability of scoring from an arbitrary point $r\in R\times R$ on the pitch, assuming the next on-ball event occurs there.
$P(C_{r})$ is the probability that the passing team will control a ball at point $r$.
$P(T_{r})$ is the probability that the next on-ball event occurs at point $r$.
Here, for simplicity, we can assume that $P(S_{r}|D), P(T_{r}|D), P(C_{r}|D)$ are independent if the parameter $\alpha=0$ in the original work implementation (Eq. (6) in  \cite{Spearman18}).
Then, the joint probability can be decomposed into a series of conditional probabilities as follows:
\vspace{-3pt}
\begin{equation}
    P(G|D)=\Sigma_{r\in R\times R}P(S_{r}|D)P(C_{r}|D)P(T_{r}|D).
\end{equation}
We show the illustrative example of OBSO in Fig \ref{fig:obso_eq}. 
$P(C_{r}|D)$ is the probability that the attacking team will control the ball at point $r$ assuming the next on-ball event occurs there, which is called the potential pitch control field (PPCF). 
$P(T_{r}|D)$ is defined as a two-dimensional Gaussian distribution with the current ball coordinates as the mean. 
The standard deviation is set to 14 m, which is the average distance of the next event \cite{Spearman18}.
$P(S_{r}|D)$ is simply calculated as a value that decreases with the distance from the goal. 
We used the grid data and computed $P(C_{r}|D)$ and $P(T_{r}|D)$ based on the code at \url{https://github.com/Friends-of-Tracking-Data-FoTD/LaurieOnTracking}.

Although $P(T_{r}|D)$ and $P(S_{r}|D)$ are simple functions, we need to explain $P(C_{r}|D)$ (PPCF) in more detail.
PPCF \cite{Spearman18} (a previous version is \cite{Spearman2017}) assumes that a player’s ability to make a controlled touch on the ball (when near the ball) can be treated as a Poisson point process. 
That is, the longer a player is near the ball without another player interfering, the more likely it becomes that they can make a controlled touch on the ball. 
The model quantifies the probability of control for each player at each location on the pitch. 
The differential equation used to compute the control probability for each player at a specified location $r$ at time $t$ is: 

\begin{align}
    \frac{\textit{d}PPCF_j(t,r,T|s,\lambda_j)}{\textit{d}T} = \left( 
    1- \sum_{k} PPCF_k(t,r,T|s,\lambda_j)
    \right) f_j(t,r,T|s)\lambda_j,
    \label{eq:ppcf}
\end{align}
where $f_j(t,r,T|s)$ represents the probability that player $j$ at time $t$ can reach location $r$ within time $T$. 
The parameter $s$ is the temporal uncertainty of player-ball intercept time (has units of $s$), which is used in $f_j(t,r,T|s)$ (we set $s=0.45$ based on \cite{Spearman2017}). 
The parameter $\lambda_j$ is the rate of control representing the inverse of the mean time (has units of $1/s$) which would take a player to make a controlled touch on the ball.
Conceptually, we consider the probability that player $j$ will be able to control the ball during time $T$ to $T+dT$ with the decay rate $f_j(t,r,T|s)\lambda_j$.
We set $PPCF_j(t,r,T|s,\lambda_j)=0$ for the attacking or defending team if the opponent team can arrive significantly before the attacking or defending team.
By integrating Eq. (\ref{eq:ppcf}) over $T$ from $0$ to $\infty$, we obtain a per-player probability of control. 
We integrate it over the players of the attacking team.
$f_j(t,r,T|s)$ is represented as a logistic function such that
\begin{align}
f_j(t,r,T|s) = \left[1+\exp\left(-\frac{T-\tau_{exp}(t,r)}{\sqrt{3}s/\pi}\right)\right]^{-1},
\end{align}
where $\tau_{exp}(t,r)$ is a expected intercept time computed from the location and velocity of the player $j$ (including other constants, see \cite{Spearman18} for the details). 
Conceptually, if $T-\tau_{exp}(t,r) \geq 0$, the player will tend to intercept the ball and a temporal uncertainty $\sqrt{3}s/\pi$ is assumed.
For the control rate parameter $\lambda_j$, higher values of $\lambda_j$ mean less time is required before the player can control the ball.
We set $\lambda_j = 4.3$ based on \cite{Spearman2017}. 

\vspace{-5pt}
\begin{figure}[h]
    \centering
    \includegraphics[scale=0.50]{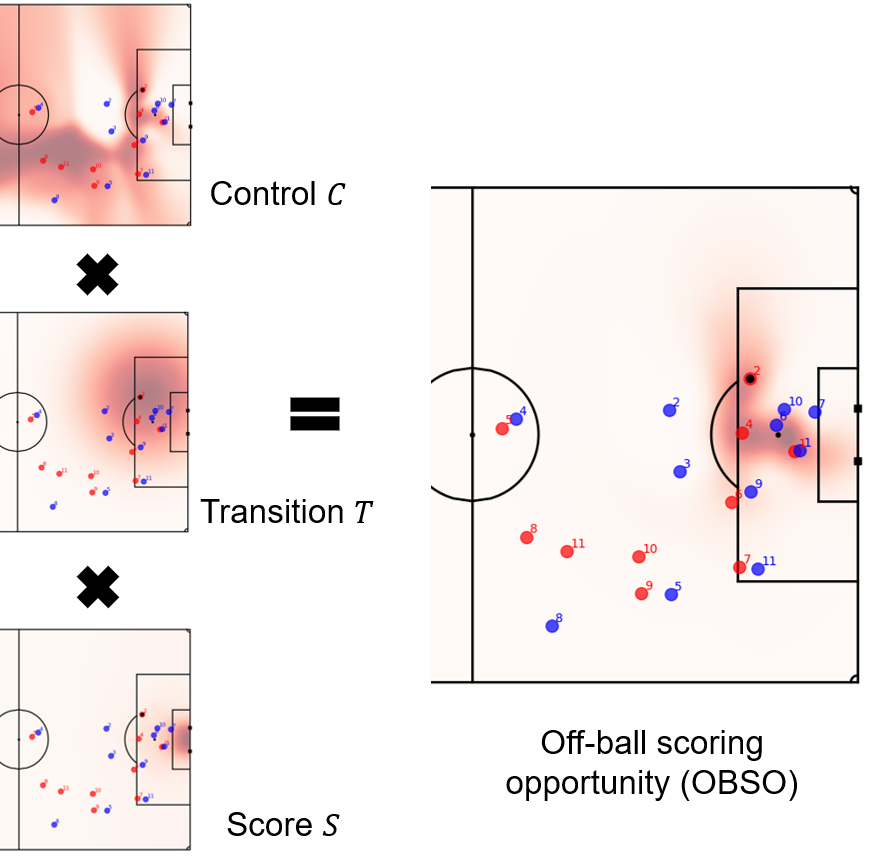}
    \caption{Off-ball scoring opportunities (OBSO) \cite{Spearman18}: an evaluation index for scoring opportunities in the off-ball state. The OBSO on the right is calculated by the joint probability of control, transition, and score probability.
    }
    \label{fig:obso_eq}
\end{figure}
\vspace{-5pt}

\section{Variational recurrent neural network \cite{Chung15}}
\label{app:vrnn}
In this section, we briefly overview recurrent neural networks (RNNs), variational autoencoders (VAEs), and variational RNNs (VRNNs).

%
From the perspective of a probabilistic generative model, an RNN models the conditional probabilities with a hidden state $h_t$ that summarizes the past history in the first $t-1$ timesteps:
\eq{
p_{\theta}(x_t | x_{< t}) = \varphi(h_{t-1}), \quad \quad h_t = f(x_t, h_{t-1}),
}
where $\varphi$ maps the hidden state to a probability distribution over states and $f$ is a deterministic function such as LSTMs 
or GRUs. 
RNNs with simple output distributions often struggle to capture highly variable and structured sequential data. Recent work in sequential generative models addresses this issue by injecting stochastic latent variables into the model and using amortized variational inference to infer latent variables from data.
VRNNs \cite{Chung15} is one of the methods using this idea and combining RNNs and VAEs. 

VAE \cite{Kingma14} is a generative model for non-sequential data that injects latent variables $z$ into the joint distribution $p_{\theta}(a, z)$ and introduces an inference network parameterized by $\phi$ to approximate the posterior $q_{\phi}(z \mid a)$. 
The learning objective is to maximize the evidence lower-bound (ELBO) of the log-likelihood with respect to the model parameters $\theta$ and $\phi$:
\eq{\mathbb{E}_{q_{\phi}(z | a)}\brcksq{\log p_{\theta}(a | z)} - D_{KL}(q_{\phi}(z \mid a) || p_{\theta}(z))}

The first term is known as the reconstruction term and can be approximated with Monte Carlo sampling. 
The second term is the Kullback-Leibler divergence between the approximate posterior and the prior, and can be evaluated analytically if both distributions are Gaussian with diagonal covariance. 
The inference model $q_{\phi}(z \mid a)$, generative model $p_{\theta}(a \mid z)$, and prior $p_{\theta}(z)$ are often implemented with neural networks.

VRNNs combine VAEs and RNNs by conditioning the VAE on a hidden state $h_t$:
\eq{
p_{\theta}(z_t | x_{<t}, z_{<t}) & = \varphi_{\text{prior}}(h_{t-1}) & & \text{(prior)} \label{eq:vrnn_prior} \\
q_{\phi}(z_t | x_{\leq t}, z_{<t}) & = \varphi_{\text{enc}}(x_t, h_{t-1}) & & \text{(inference)} \\
p_{\theta}(x_t | z_{\leq t}, x_{<t}) & = \varphi_{\text{dec}}(z_t, h_{t-1}) & & \text{(generation)} \\
h_t & = f(x_t, z_t, h_{t-1}). & & \text{(recurrence)}
\label{eq:vrnn_state}
}
VRNNs are also trained by maximizing the ELBO, which can be interpreted as the sum of VAE ELBOs over each timestep of the sequence:
\eq{
\mathbb{E}_{q_{\phi}(z_{\leq T} \mid x_{\leq T})} & \Bigg[ \sum_{t=1}^T \log p_{\theta}(x_t \mid z_{\leq T}, x_{<t}) \\
& - D_{KL} \Big( q_{\phi}(z_t \mid x_{\leq T}, z_{<t}) || p_{\theta}(z_t \mid x_{<t}, z_{<t}) \Big) \Bigg] \nonumber
}
Note that the prior distribution of latent variable $z_t$ depends on the history of states and latent variables (Eq. (\ref{eq:vrnn_prior})). 

\section{Graph variational recurrent neural network \cite{Yeh2019}}
\label{app:GVRNN}
Here, we briefly describe VRNN, GNN, and GVRNN.

In general, RNNs with simple output distributions often struggle to capture highly variable and structured sequential data (e.g., multimodal behaviors) \cite{Zhan19}. 
Recent work in sequential generative models addressed this issue by injecting stochastic latent variables into the model and optimization using amortized variational inference to learn the latent variables (e.g., \cite{Chung15,Fraccaro16,Goyal17}). %
Among these methods, a variational RNN (VRNN) \cite{Chung15} has been widely used in base models for multi-agent trajectories \cite{Yeh2019,Zhan19,fujii2020policy} with unknown governing equations.
A VRNN is essentially a variational autoencoder (VAE) conditioned on the hidden state of an RNN and is trained by maximizing the (sequential) evidence lower-bound (ELBO), described in Appendix A. 

Next, we overview a graph neural network (GNN) based on \cite{Kipf18}.
Let $v_k$ be a feature vector for each node $k$ of $K$ agents. 
Next, a feature vector for each edge $e_{(k,j)}$ is computed based on the nodes to which it is connected. 
The edge feature vectors are sent as ``messages'' to each of the connected nodes to compute their new output state $o_k$.
Formally, a single round of message passing operations of a graph net is characterized below:
\eq{
v\rightarrow e: e_{(k,j)} &= ~f_e([v_k,v_j]), \\
e\rightarrow v: ~~~~~~o_i &= ~f_v\left(\sum_{j\in N(k)} e_{(k,j)}\right),
\label{eq:message} 
}
where $N(k)$ is the set of neighbors of node $k$, and $f_e$ and $f_v$ are neural networks.
In summary, a GNN takes in feature vectors $v_{1:K}$ and outputs a vector for each node $o_{1:K}$, i.e., $o_{1:K} = {\mathrm{GNN}}(v_{1:K})$. 
The operations of the GNN satisfy the permutation equivariance property as the edge construction is symmetric between pairs of nodes and the summation operator ignores the ordering of the edges \cite{zaheer2017deep}.

Next, we describe GVRNN \cite{Yeh2019}, which models the interactions between them at each step using GNNs. 
Let $x_{\leq T} = \{ x_1, \dots, x_T \}$ denote a sequence of locations. 
In this paper, GVRNN update equations are as follows:
\eq{
p_{\theta}(z_t | x_{<t}, z_{<t}) & = \prod_k \mathcal{N}(z_{t,k}|\mu^{\mathrm{pri}}_{t,k},(\sigma^{\mathrm{pri}}_{t,k})^2), \\
q_{\phi}(z_t | x_{\leq t}, z_{<t}) & = \prod_k \mathcal{N}(z_{t,k}|\mu^{\mathrm{enc}}_{t,k},(\sigma^{\mathrm{enc}}_{t,k})^2),\\
p_{\theta}(x_t | z_{\leq t}, x_{<t}) & = \prod_k \mathcal{N}(z_{t,k}|\mu^{\mathrm{dec}}_{t,k},(\sigma^{\mathrm{dec}}_{t,k})^2),\\
h_{t,k} & = f_{rnn}(x_{t,k}, z_{t,k}, h_{t-1,k}). 
\label{eq:gvrnn_state}
}
where $h_t$ and $z_t$ are deterministic and stochastic latent variables.
$p_{\theta}(x_t \mid z_{\leq t}, x_{<t})$, $q_{\phi}(z_t \mid x_{\leq t}, z_{<t})$, and $p_{\theta}(z_t \mid x_{<t}, z_{<t})$ are generative model, the approximate posterior or inference model, and the prior model, respectively.
$\mathcal{N}(\cdot|\mu,\sigma^2)$ denotes a multivariate normal distribution with mean $\mu$ and covariance matrix diag($\sigma^2$), and
\eq{
[\mu^{\mathrm{pri}}_{t,1:K},\sigma^{\mathrm{pri}}_{t,1:K}] & = {\mathrm{GNN_{pri}}}(h_{t-1,1:K}),\\
[\mu^{\mathrm{enc}}_{t,1:K},\sigma^{\mathrm{enc}}_{t,1:K}] & = {\mathrm{GNN_{enc}}}([x_{t,1:K},h_{t-1,1:K}]), \\
[\mu^{\mathrm{dec}}_{t,1:K},\sigma^{\mathrm{dec}}_{t,1:K}] & = {\mathrm{GNN_{dec}}}([z_{t,1:K},h_{t-1,1:K}]).
}
The prior network ${\mathrm{GNN_{pri}}}$, encoder ${\mathrm{GNN_{enc}}}$, and decoder ${\mathrm{GNN_{dec}}}$ are GNNs with learnable parameters $\phi$ and $\theta$.
Here we used the mean value $\mu^{\mathrm{dec}}_{t+1,1:K}$ as input variables $\hat{x}^{l'}_{t+1}$ in the following theory-based computation.
GVRNN is trained by maximizing the sequential ELBO in a similar way to VRNN as described in Appendix \ref{app:vrnn}.

\section{Validation results of trajectory prediction model}
\label{app:prediction_results}

To verify the accuracy of the trajectory prediction model, We compared our approach with two baselines: VRNN \cite{Chung15} and RNN (RNN$+$Gauss) implemented using a gated recurrent unit (GRU) \cite{Cho14} and a decoder with Gaussian distribution for prediction \cite{Becker18}. 

For the MAE in the trajectory, to compare the various methods and time lengths, we first performed the Kruskal-Wallis test.
As the post-hoc comparison, since we are interested in the differences from GVRNN (VRNN and RNN for four time lengths) and time lengths in GVRNN (4 and 6, 6 and 8, and 8 and 10 s), we performed the Wilcoxon rank sum test with Bonferroni correction such that the p-value was multiplied by 11 ($4 \times 2 + 3$).

We show the results of the trajectory prediction model for computing C-OBSO.
The endpoint errors (MAE) of the three players are shown in Table \ref{tab:prediction_performance}. 
In the statistical evaluation, there were significant differences in all classification performance and tasks ($p < 10^{-10}$) using the Kruskal-Wallis test.
In the following evaluations, we indicate the post-hoc comparison results.
The trajectory prediction model used in C-OBSO (GVRNN) shows a lower prediction error than other models ($ps < 10^{-10}$).

In GVRNN, longer predictions show larger prediction errors ($ps < 10^{-10}$) except for the difference between 8 s and 10 s ($p > 0.05$).
Note that we verified the existing GVRNN \cite{Yeh2019} performance, which uses a centralized optimization whereas VRNN and RNN$+$Gauss use the decentralized optimization (for each player). 
Since the 4 s prediction of GVRNN achieved a low the MAE of less than 0.7 m, the GVRNN trajectory prediction of 4 s was used in the next C-OBSO.

\ifarxiv
\begin{table}[h]
    \caption{Trajectory prediction endpoint errors (MAE and standard error) in three methods.
    }
    \centering
    \scalebox{0.85}{
    \begin{tabular}{c|cccc}
        \hline
         & 4 s  & 6 s & 8 s & 10 s\\ 
        \hline
        GVRNN & \textbf{0.608} $\pm$ \textbf{0.014}& \textbf{0.867} $\pm$ \textbf{0.022}& \textbf{1.701} $\pm$ \textbf{0.045} & \textbf{1.606} $\pm$ \textbf{0.042}\\
        \hline
        VRNN & 5.952 $\pm$ 0.118& 7.767 $\pm$ 0.160& 9.127 $\pm$ 0.188 & 10.168 $\pm$ 0.225\\
        RNN+Gauss & 9.101 $\pm$ 0.144 & 11.396 $\pm$ 0.202& 13.312 $\pm$ 0.245 & 15.327 $ \pm$ 0.302\\
    \end{tabular}
    }
    \label{tab:prediction_performance}
\end{table}
\fi


\section{C-OBSO and OBSO results without the potential score model}
\label{app:withoutpotential}
To investigate the effect of the potential model on the C-OBSO and OBSO computations, we also computed C-OBSO and OBSO results without the potential score model.
Results shown in Fig. \ref{fig:withoutpotential} were similar to those with the potential model, but there were no significant correlations between the C-OBSO and salary ($\rho = 0.38, p = 0.08$) and between the OBSO and salary ($\rho = -0.18, p = 0.26$).

\begin{figure}[h]
    \centering
    \includegraphics[scale=0.45]{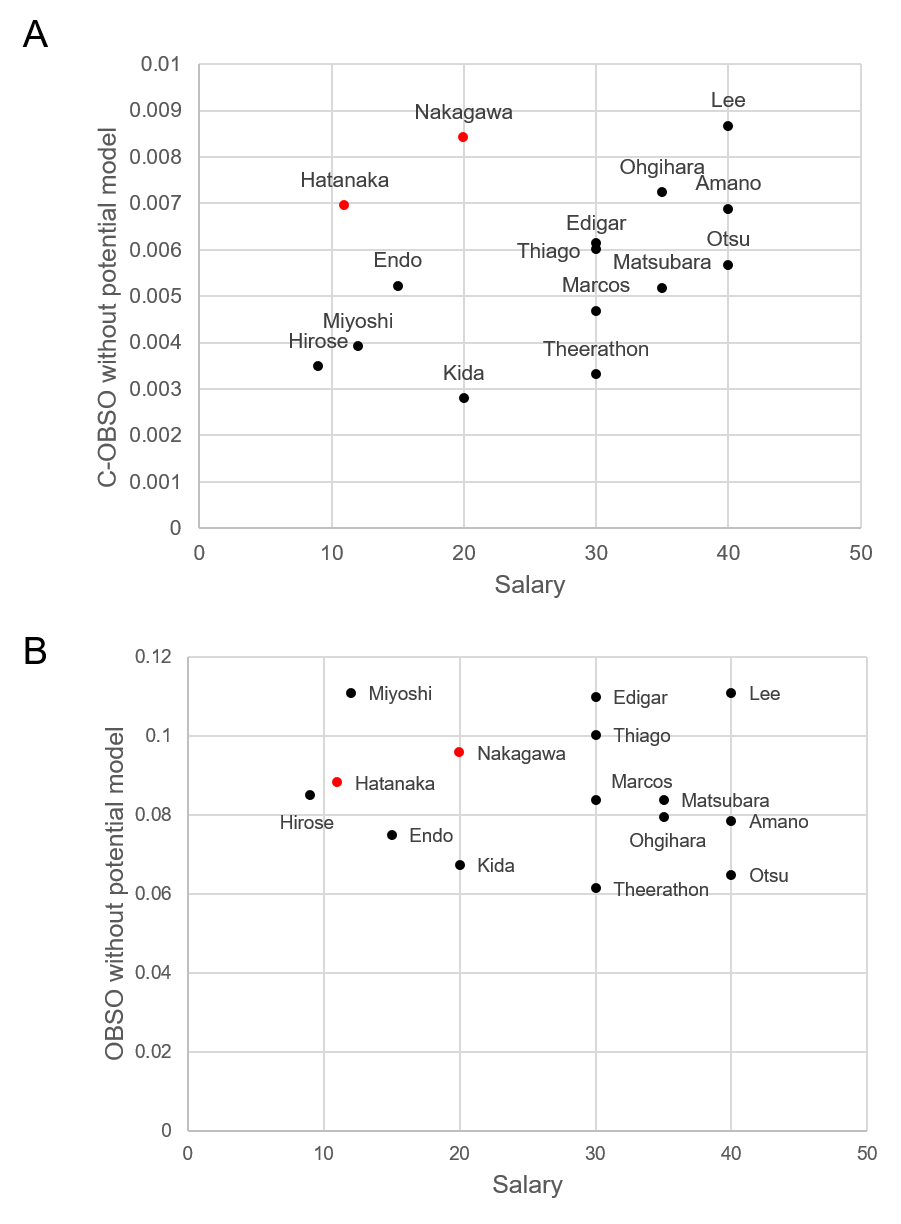}
    \vspace{-10pt}
    \caption{Relationship between indicators without the potential score model and annual salary in a team. 
    (A) Relationship between C-OBSO without the potential model and the salary. 
    (B) Relationship between OBSO \cite{Spearman18} without the potential model and the salary. 
    Configurations are same as Fig. \ref{fig:salary_cobso}. 
    }
    \label{fig:withoutpotential}
\end{figure}

\section{Relationship between rating, C-OBSO, and goal}
\label{app:rating}
We additionally analyzed the relationship between the game rating by experts, C-OBSO, and the number of goals. 
First, we show the relationship between C-OBSO and the rating by experts of the top seven scorers in Table \ref{tab:cobso_scoreplayer}.
We analyzed seven players who played seven games or more and had two related scoring opportunities or more (for C-OBSO). 
There were no significant correlations between them for all players ($\rho s < 0.190,~ ps > 0.05$) except for Nakagawa. 

Next, we also show the relationship between the goals of the top seven scorers and the ratings by experts in Table \ref{tab:rating_scoreplayer}. 
We analyzed the games in which each player scored two points or more.
There were significant correlations between them for all players ($\rho s > 0.516,~ ps < 0.018$).

\begin{table*}[h!]
    \caption{Relationship between C-OBSO and the rating by experts of the top seven creator of scoring opportunities for teammate for each game. We analyzed seven players who played seven games or more and had two related scoring opportunities or more (for C-OBSO). The OBSO values were different from Fig. \ref{fig:salary_cobso} because the values in this table were computed by the mean and standard deviation of the mean value of each game.
    }
    \centering
    \scalebox{1}{
    \begin{tabular}{c|c|c|c|c|c|c}
        \hline
Name & Position & No. of games & Rating   & C-OBSO   & $\rho$ & $p$ \\

\hline
Nakagawa & FW & 17 & 6.18 $\pm$ 0.64 & 0.0043 $\pm$ 0.00624 & 0.751 & 0.0003 \\
Marcos & FW & 14 & 6.05 $\pm$ 0.60 & 0.0032 $\pm$ 0.00324 & 0.272 & 0.1738 \\
Edigar & FW & 10 & 6.00 $\pm$ 0.71 & 0.0038 $\pm$ 0.00291 & -0.371 & 0.1454 \\
Endo & MF & 7 & 5.86 $\pm$ 0.56 & 0.0020 $\pm$ 0.00248 & -0.418 & 0.1751 \\
Amano & MF & 7 & 5.86 $\pm$ 0.38 & 0.0086 $\pm$ 0.00617 & -0.116 & 0.4024 \\
Ohgihara & MF & 7 & 6.00 $\pm$ 0.29 & 0.0123 $\pm$ 0.01073 & -0.134 & 0.3876 \\
Matsubara & DF & 7 & 6.14 $\pm$ 0.63 & 0.0079 $\pm$ 0.01335 & 0.189 & 0.3426 \\

    \end{tabular}
    }
    \label{tab:cobso_scoreplayer}
\end{table*}

\begin{table*}[h!]
    \caption{Relationship between the goals and the rating by experts of the top seven scorers. We analyzed the games in which each player scored two points or more. The ratings were different from Table \ref{tab:cobso_scoreplayer} because of the different data selection criteria.
    }
    \centering
    \scalebox{1}{
    \begin{tabular}{c|c|c|c|c|c|c}
        \hline
         Name & Position & No. of games & No. of goals & Rating   & $\rho$ & $p$ \\
 
        \hline
Nakagawa & FW & 33 & 15 & 6.33 $\pm$ 0.79 & 0.630 & 4.33E-05 \\
Marcos & FW & 33 & 15 & 6.09 $\pm$ 0.92 & 0.709 & 1.98E-06 \\
Edigar & FW & 16 & 11 & 6.14 $\pm$ 0.75 & 0.912 & 4.40E-07 \\
Erik & FW & 12 & 8 & 6.13 $\pm$ 0.77 & 0.599 & 1.97E-02 \\
Endo & MF & 29 & 7 & 5.88 $\pm$ 0.61 & 0.613 & 2.03E-04 \\
Theerathon & DF & 25 & 3 & 5.88 $\pm$ 0.62 & 0.517 & 4.09E-03 \\
Miyoshi & MF & 16 & 3 & 6.03 $\pm$ 0.64 & 0.529 & 1.75E-02 \\

    \end{tabular}
    }
    \label{tab:rating_scoreplayer}
\end{table*}
\fi

\end{document}